\documentclass[runningheads]{llncs}
\usepackage{graphicx}
\usepackage{amsmath,amssymb} 
\usepackage{color}

\usepackage{subcaption}
\begin{document}
\title{Embarrassingly Simple Model \\ for Early Action Proposal} 

\titlerunning{Embarrassingly Simple Model for Early Action Proposal}
%
\author{M. Baptista-R\'ios, R. J. L\'opez-Sastre, \\ F. J. Acevedo-Rodr\'iguez and S. Maldonado-Basc\'on.}
\institute{GRAM, Department of Signal Theory and Communications, University of Alcal\'a
\\
\textbf{Published in the Anticipating Human Behavior Workshop, ECCV 2018}}

%
\authorrunning{M. Baptista-R\'ios et al.}
%

%
\maketitle              
\vspace{-0.9cm}
\begin{abstract}
Early action proposal consists in generating high quality candidate temporal segments that are likely to contain an action in a video stream, as soon as they happen. Many sophisticated approaches have been proposed for the action proposal problem but from the off-line perspective. On the contrary, we focus on the \emph{on-line} version of the problem, proposing a simple classifier-based model, using standard 3D CNNs, that performs significantly better than the state of the art.
\vspace{-0.2cm}
\end{abstract}
\vspace{-0.6cm}
\textbf{Introduction}. In this work, we introduce the novel problem of Early Action Proposal (EAP). Unlike traditional off-line activity proposal approaches \cite{sparseprop,Escorcia2016,scnn_cvpr16,Gao_2017_ICCV,chao:cvpr2018}, we move towards the on-line version of the problem, where the goal is to generate high quality action candidate temporal segments in a video stream, but as soon as they happen. This novel \emph{early} setting can be useful in many practical applications, where the video arrives in an on-line fashion, such as for robotics, or video surveillance cameras. Moreover, we show that the sophisticated off-line solutions \cite{sparseprop,Escorcia2016,scnn_cvpr16,Gao_2017_ICCV,chao:cvpr2018}, which define the current state-of-the-art, offer a poor performance for the EAP problem, mainly because they assume a more simplified setup, where the whole video is always available to produce the proposals.

\begin{figure}
	\centering
	\includegraphics[width=0.8\textwidth]{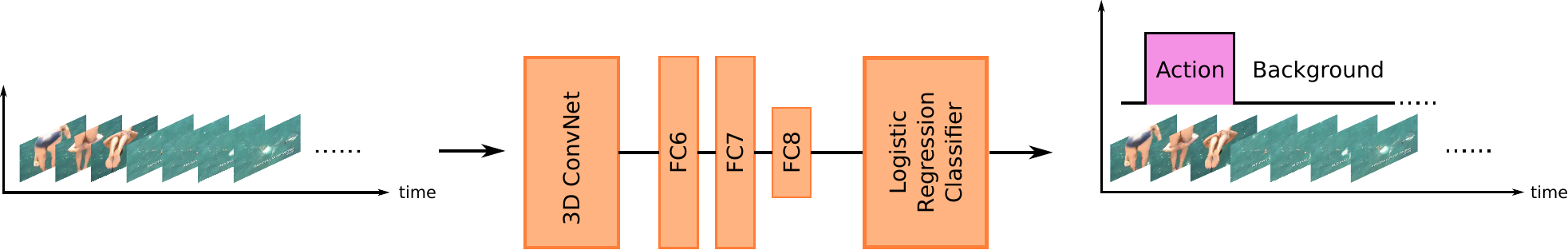}
	\caption{Proposed C3D-EAP model. We use a logistic regression classifier, with C3D features, trained to discriminate action from background.}
	\label{fig:graphical_abstract}
\end{figure}

\textbf{Model description}. As it is shown in Figure \ref{fig:graphical_abstract}, for EAP, the action proposal must be generated on-line. This requires identifying whether the action is taking place or not, directly from the video stream, hence working with partial observations of the actions, and ideally with a minimal latency, which is in contrast to the complex sampling mechanisms of most of the off-line models. In addition, an EAP solution must correctly discriminate the action from the background frames, the latter being more frequent.
\vspace{-0.8cm}

\vspace{-0.7cm}
We propose a simple solution, which is based on the 3D CNNs (C3D) \cite{c3d-dutran}. We name our model as C3D-EAP. Technically, it consists of learning a C3D network to discriminate between action and background. Then, for each set of frames of a given test video, our network indicates whether they are action or background. With this output (see Figure \ref{fig:graphical_abstract}), we can build the action proposals in an on-line fashion. In order to assign a score for each of our early proposals, we use the mean of the scores, provided by the logistic regression classifier, for the set of evaluated frames of the proposal.

\textbf{Experiments}. For the experimental validation of the EAP problem, we use the untrimmed videos from the THUMOS'14 dataset \cite{THUMOS14}. We report results on the 213 test videos, using the validation set for learning our approach. We compare our work (C3D-EAP) with those state-of-the-art approaches whose authors provide results, \emph{i.e.} Sparse-Prop \cite{sparseprop}, DAPs \cite{Escorcia2016}, and the recent TURN-TAP \cite{Gao_2017_ICCV}. 

\begin{figure}
	\centering
	\begin{subfigure}{.5\textwidth}
		\centering
		\includegraphics[width=.7\linewidth]{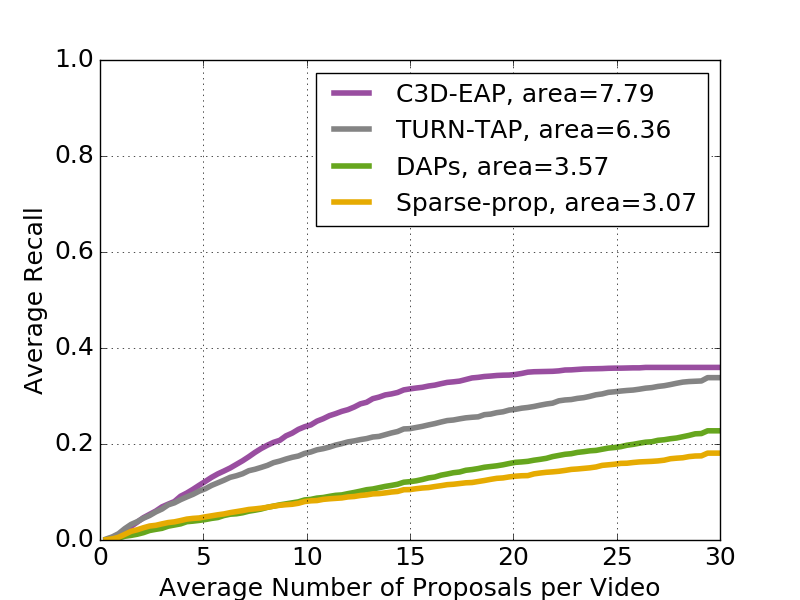}
		\caption{AR-AN with 0.5 tIoU threshold.}
		\label{subfig:ar-an}
	\end{subfigure}%
	\begin{subfigure}{.5\textwidth}
		\centering
		\includegraphics[width=.7\linewidth]{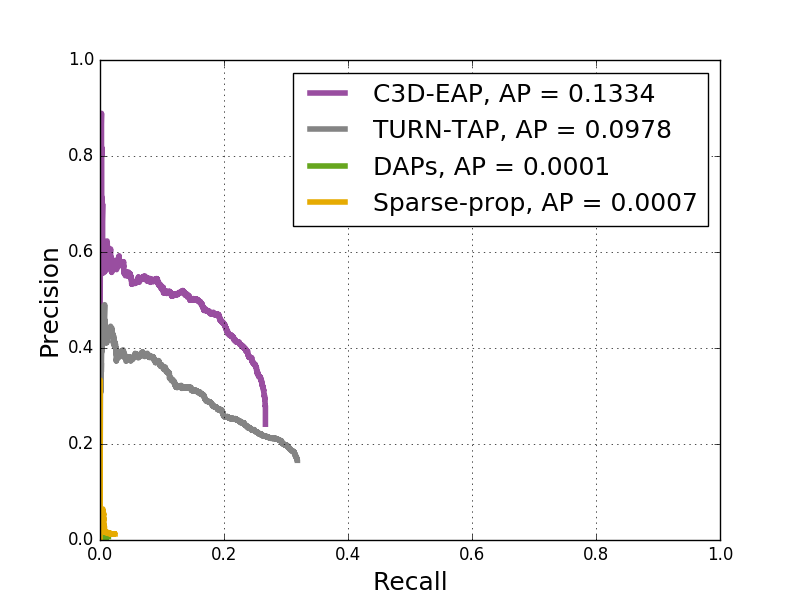}
		\caption{P-R with 0.5 tIoU threshold.}
		\label{subfig:p-r}
	\end{subfigure}
	\caption{Comparison with the state-of-the-art on THUMOS'14 \cite{THUMOS14} dataset.}
	\label{fig:metrics}
\end{figure}

As for the evaluation, we use the standard metric Average Recall at different Average Number of Proposals per Video (AR-AN), used by all the off-line state-of-the-art models \cite{sparseprop,Escorcia2016,scnn_cvpr16,Gao_2017_ICCV,chao:cvpr2018}. However, for the novel EAP problem it is important to also evaluate whether the proposals are likely to include
the action of interest, ideally achieving high recall with few
proposals. Therefore, as it is shown in Figure \ref{subfig:ar-an}, we limit the number of proposals to 30 per video\footnote{In the THUMOS'14 test set, each video contains 15 action instances on average.}, during the evaluation. Furthermore, to measure the quality of the proposals for the EAP problem, we also advocate that it is important to recover the well-known Precision-Recall (PR) curve as an evaluation metric (see Figure \ref{subfig:p-r}). 

\vspace{-0.7cm}

\begin{figure}[h]
	\centering
	\begin{subfigure}{.33\textwidth}
		\centering
		\includegraphics[width=\linewidth]{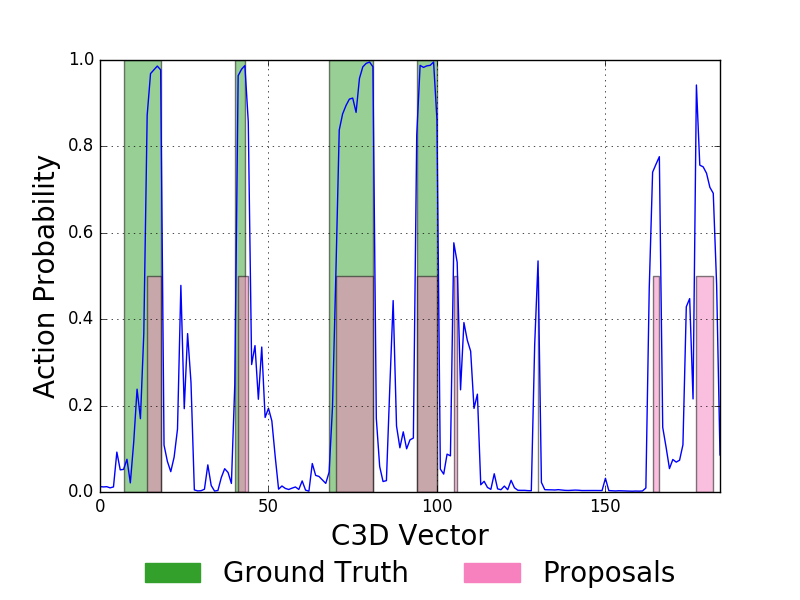}
		\label{subfig:73}
	\end{subfigure}%
	\begin{subfigure}{.33\textwidth}
		\centering
		\includegraphics[width=\linewidth]{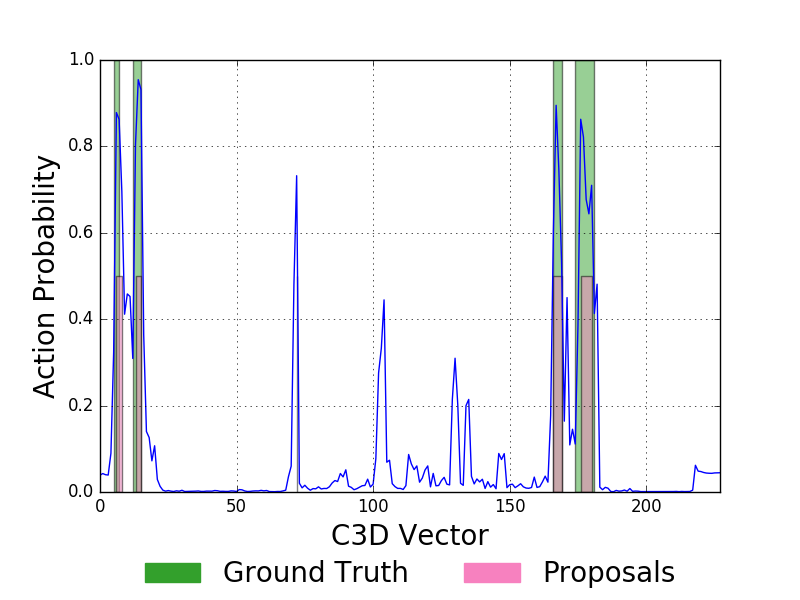}
		\label{subfig:374}
	\end{subfigure}
	\begin{subfigure}{.33\textwidth}
		\centering
		\includegraphics[width=\linewidth]{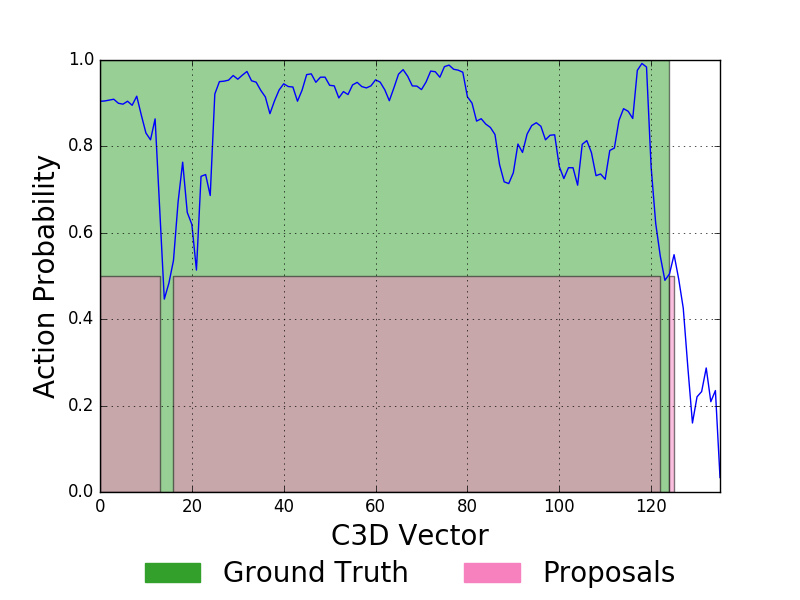}
		\label{subfig:504}
	\end{subfigure}
	\vspace{-0.7cm}
	\caption{Qualitative results for our C3D-EAP approach.}
	\label{fig:qualitative}
\end{figure}

\vspace{-0.7cm}
Figure \ref{fig:metrics} shows that our embarrassingly simple model clearly outperforms the state-of-the-art approaches for both metrics. While those approaches generate many proposals in an off-line fashion, our solution is able to work on-line, generating fewer proposals, that are also more accurate. This can be seen in the qualitative examples presented in Figure \ref{fig:qualitative}.

\bibliographystyle{splncs04}
\bibliography{egbib}

\end{document}